\documentclass{llncs}
\usepackage{graphicx}
\usepackage{listings}
\usepackage{color}
\usepackage{algorithm}
\usepackage{algorithmic}
\usepackage{subfigure}
\usepackage{mathrsfs}
\usepackage{graphicx}
\usepackage{comment}
\usepackage{amsmath,amssymb} 
\usepackage{color}
 
\definecolor{codegreen}{rgb}{0,0.6,0}
\definecolor{codegray}{rgb}{0.5,0.5,0.5}
\definecolor{codepurple}{rgb}{0.58,0,0.82}
\definecolor{backcolour}{rgb}{0.95,0.95,0.92}
 
\lstdefinestyle{mystyle}{
    backgroundcolor=\color{backcolour},   
    commentstyle=\color{codegreen},
    keywordstyle=\color{magenta},
    numberstyle=\tiny\color{codegray},
    stringstyle=\color{codepurple},
    basicstyle=\footnotesize,
    breakatwhitespace=false,         
    breaklines=true,                 
    captionpos=b,                    
    keepspaces=true,                 
    numbers=left,                    
    numbersep=5pt,                  
    showspaces=false,                
    showstringspaces=false,
    showtabs=false,                  
    tabsize=2
}
 
\begin{document}

\title{Fine-Grained Visual Classification via Progressive Multi-Granularity Training of Jigsaw Patches}
\author{Ruoyi Du\inst{1} \and
Dongliang Chang\inst{1} \and
Ayan Kumar Bhunia\inst{2} \and
Jiyang Xie\inst{1} \and
Yi-Zhe Song\inst{2} \and
Zhanyu Ma\inst{1} \and
Jun Guo\inst{1}}

\institute{Beijing University of Posts and Telecommunications
\email{\{beiyoudry,changdongliang,xiejiyang2013,mazhanyu,guojun\}@bupt.edu.cn}\\
\and
University of Surrey\\
\email{\{a.bhunia,y.song\}@surrey.ac.uk}}

\maketitle              

\begin{abstract}
Fine-grained visual classiﬁcation (FGVC) is much more challenging than traditional classiﬁcation tasks due to the inherently subtle intra-class object variations. Recent works mainly tackle this problem by focusing on how to locate the most discriminative parts, more complementary parts, and parts of various granularities. However, less effort has been placed to which granularities are the most discriminative and how to fuse information cross multi-granularity. In this work, we propose a novel framework for fine-grained visual classiﬁcation to tackle these problems. In particular, we propose: (i) a novel progressive training strategy that adds new layers in each training step to exploit information based on the smaller granularity information found at the last step and the previous stage. (ii) a simple jigsaw puzzle generator to form images contain information of different granularity levels. We obtain state-of-the-art performances on several standard FGVC benchmark datasets, where the proposed method consistently outperforms existing methods or delivers competitive results. The code will be available at \textit{https://github.com/RuoyiDu/PMG-Progressive-Multi-Granularity-Training}
\end{abstract}

\newpage
\section{Introduction}

Fine-grained visual classification (FGVC) aims at identifying sub-classes of a given object category, {\em e.g.}, different species of birds, and models of cars and aircrafts. It is a much more challenging problem than traditional classification due to the inherently subtle intra-class object variations amongst sub-categories. Most effective solutions to date rely on extracting fine-grained feature representations at local discriminative regions, either by explicitly detecting semantic parts \cite{fu2017look,zheng2017learning,yang2018learning,ge2019weakly,zhang2019learning} or implicitly via saliency localization \cite{wang2018learning,dubey2018pairwise,chen2019destruction,luo2019cross}. It follows that such locally discriminative features are collectively fused to perform final classification.

Early work mostly finds discriminative regions with the assistance of manual annotations \cite{berg2013poof,lei2016fast,xie2013hierarchical,zhang2014part,huang2016part}. However, human annotations are difficult to obtain, and can often be error-prone resulting in performance degradations \cite{zheng2017learning}. Research focus has consequently shifted to training models in a weakly-supervised manner given only category labels \cite{zheng2017learning,yang2018learning,wang2018learning,chen2019destruction}. Success behind these models can be largely attributed to being able to locate more discriminative local regions for downstream classification. However little or no effort has been made towards (i) at which granularities are these local regions most discriminative, e.g., head or beak of a bird, and (ii) how can information across different granularities be fused together to classification accuracy, e.g., can do head and beak work together. 

Information cross various granularities is however helpful for avoiding the effect of large intra-class variations. For example, experts sometimes need to identify a bird using \textit{both} the overall structure of a bird's head, and finer details such as the shape of its beak. That is, it is often not sufficient to identify discriminative parts, but also how these parts interact amongst each other in a complementary manner. Very recent research has focused on the ``zooming-in'' factor \cite{fu2017look,zhang2019learning}, i.e., not just identifying parts, but also focusing on the truly discriminative regions within each part (e.g., the beak, more than the head). Yet these methods mostly focuses on a few parts and ignores others as zooming in beyond simple fusion. More importantly, they do not consider how features from different zoomed-in parts can be fused together in a synergistic manner. Different to these approaches, we further argue that, one not only needs to identify parts and their most discriminative granularities, but meanwhile how parts at different granularities can be effectively merged.



\begin{figure}[ht]
\centering
\includegraphics[width=12cm]{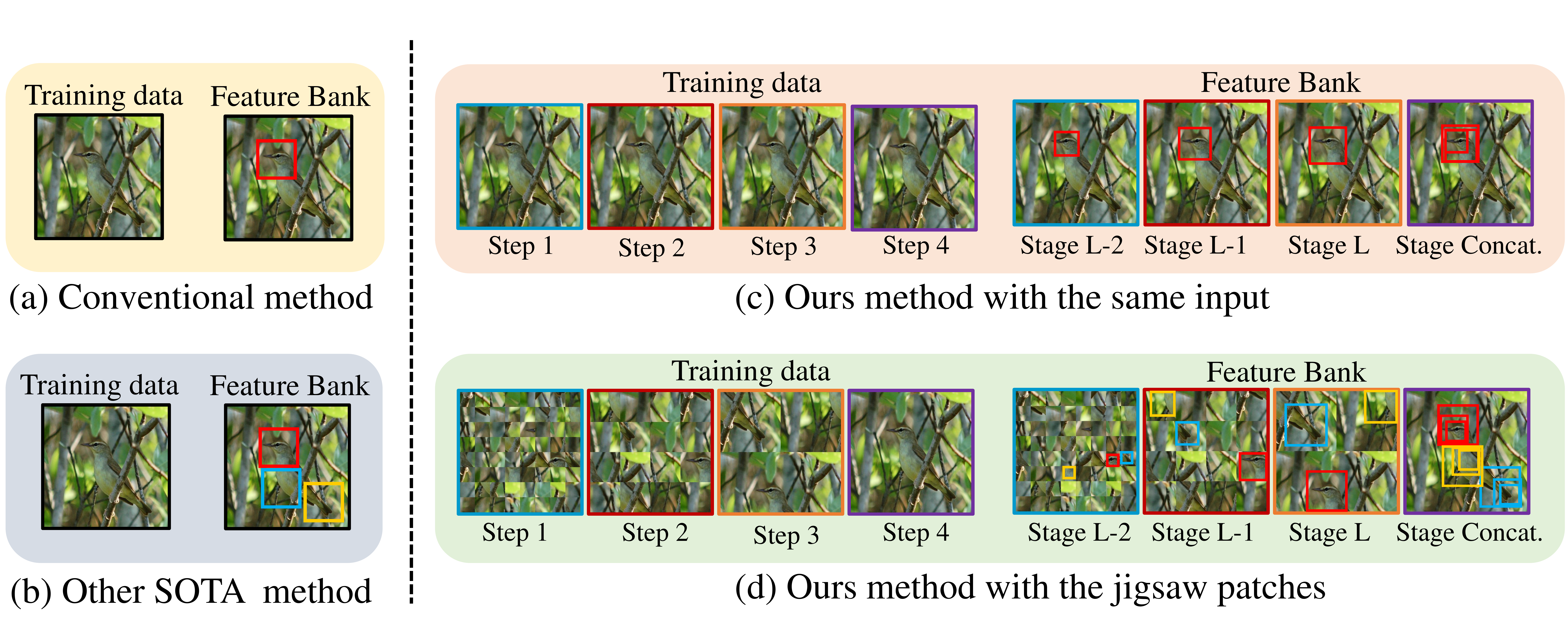}
  \caption{Illustration of features learned by general methods (a and b) and our proposed method (c and d). (a) Traditional convolution neural networks trained with cross entropy (CE) loss tend to find the most discriminative parts. (b) Other state-of-the-art methods focus on how to find more discriminative parts. (c) Our proposed progressive training (Here we use last three stages for explanation.) gradually locates discriminative information from low stages to deep stage. And features extracted from all trained stages are concatenated together to ensure complementary relationships are fully explored, which is represented by ``Stage Concat.'' (d) With assistance of jigsaw puzzle generator the granularity of parts learned at each step are restricted inside patches.}
  \label{fig:split_example}
\end{figure}

In this paper, we take an alternative stance towards fine-grained classification. We do not explicitly, nor implicitly attempt to mine fine-grained feature representations from parts (or their zoomed-in versions). Instead, we approach the problem with the hypothesis that the fine-grained discriminative information lies \textit{naturally} within different visual granularities -- it is all about encouraging the network to learn at different granularities and simultaneously fusing multi-granularity features together. This can be better explained by Figure \ref{fig:split_example}.

More specifically, we propose a consolidated framework that accommodates part granularity learning and cross-granularity feature fusion simultaneously. This is achieved through two components that  work synergistically with each other: (i) a progressive training strategy that effectively fuses features from different granularities, and (ii) a random jigsaw patch generator that encourages the network to learn features at specific granularities. Note that we refrain from using ``scale'' since we do not apply Gaussian blur filters on image patches, rather we evenly divide and shuffle image patches to form different granularity levels.


As the first contribution, we propose a multi-granularity progressive training framework to learn the complementary information across different image granularities. This differs significantly to prior art where parts are first detected, and later fused in an ad-hoc manner. Our progressive framework works in steps during training, where at each step the training focuses on cultivating granularity-specific information with a corresponding stage of the network. We start with finer granularities which are more stable, gradually move onto coarser ones, which avoids the confusion made by large intra-class variations that appear in large regions. On its own, this is akin to a ``zooming out'' operation, where the network would focus on a local region, then zoom out a larger patch surrounding this local region, and finish when we reach the whole image. More specifically, when each training step ends, the parameters trained at the current step will pass onto the next  training step as its parameter initialization. This passing operation essentially enables the network to mine information of larger granularity based on the region learned in its previous training step. Features extracted from all stages are concatenated only at the last step to further ensure complementary relationships are fully explored.

However, applying progressive training naively would not benefit fine-grained feature learning. This is because the mulit-granularity information learned via progressive training may tend to focus on the similar region. As the second contribution, we tackle this problem by introducing a jigsaw puzzle generator to form different granularity levels at each training step, and only the last step is still trained with original images. This effectively encourage the model to operate on patch-level, where patch sizes are specific to a particular granularity. It essentially forces each stages of the network to focus on local patches other than holistically across the entire image, therefore learning information specific to a given granularity level. This effect in demonstrated in Figure \ref{fig:split_example}. Note that, the very recent work of \cite{chen2019destruction} first adopted a jigsaw solver to solve for fine-grained classification. We differ significantly in that we do not employ jigsaw solver as part of feature learning. Instead, we simply generate jigsaw patches randomly as means of introducing different object parts levels to assist progressive training.

Main contributions of this paper can be summarized as follows:

  \begin{enumerate}
  \item We propose a novel progressive training strategy to solve for fine-grained classification. It operates in different training steps, and at each step fuses data from previous levels of granularity, ultimately cultivating the inherent complementary properties across different granularities for fine-grained feature learning.
    
   \item We adapt a simple yet effective jigsaw puzzle generator to form different levels of granularity. This allows the network to focus on different ``scales'' of features as per prior work.
    
    \item The proposed Progressive Multi-Granularity (PMG) training framework obtains state-of-the-art or competitive performances on all three standard FGVC benchmark datasets.
    \end{enumerate}

\section{Related Work}

\subsection{Fine-Grained Classification}

Benefiting from the recent development of neural networks \emph{e.g.}, VGG \cite{simonyan2014very} and ResNet \cite{he2016deep}, the feature extraction capabilities of the neural networks have been significantly improved. Recent studies about FGVC have moved from strongly-supervised scenario with extra annotations \emph{e.g.}, bounding box~\cite{berg2013poof,lei2016fast,xie2013hierarchical,zhang2014part,huang2016part} to weakly-supervised conditions with only category label~\cite{fu2017look,zheng2017learning,yang2018learning,ge2019weakly,zhang2019learning}. 

In the weakly supervised configuration, recent studies mainly focus on locating the most discriminative parts, more complementary parts, and parts of various granularities. However, few of them consider that how to fuse information from these discriminative parts together better, and the current fusion techniques can be roughly divided into two categories. (i) The first technique conducts predictions based on different parts and then directly combines their probabilities together \cite{zhang2019learning}. Zhang {\em et al.} \cite{zhang2019learning} trained several networks focusing on features of different granularities to produce diverse prediction distribution, and then weighting their results before combine them together. (ii) Some other methods concatenate features extracted from different parts together for next prediction  \cite{zheng2017learning,fu2017look,ge2019weakly,yang2018learning}. Fu {\em et al.} found region detection and ﬁne-grained feature learning can reinforce each other, and built a series of networks which find discriminative regions for the next network as they conducting predictions. With similar motivation, Zheng {\em et al.} \cite{zheng2017learning} jointly learned part proposals and the feature representations on each part, and located various discriminative parts before prediction. Both of them train a fully-connected fusion layer to fuse features extracted from different parts. Ge {\em et al.} \cite{ge2019weakly} went one step further by fusing features from complementary object parts with two LSTMs stacked together.

Fusion features from different parts is still a challenge problem but few efforts have been made for it. In this work, we try to address it based on the Intrinsic characteristics of fine-grained objects: although with large intra-class variation, the subtle details show stability at local regions. Hence, instead of locating the discriminative first, we guide the network to learn features from small granularity to large granularity progressively.


\subsection{Image Splitting Operation}
Splitting an image into pieces with the same size has been utilized for various goals in previous works. Among them, one typical solution is to solve the jigsaw puzzle \cite{cho2010probabilistic,son2014solving}. It can also go one step further by adopting the jigsaw puzzle solution as the initialization to a weakly-supervised network, which leads to better transformation performance~\cite{wei2019iterative}. This method helps the network exploit the spatial relationship of images. In one-shot learning, image splitting operation is used for augmentation, which split two image and exchange some patches of them to generate new training ones \cite{chen2019image}. In more recent research, DCL \cite{chen2019destruction} first adopt image splitting operation for FGVC by destructing the global structure to emphasis local details and reconstructing the images to learn semantic correlation among local regions. However, it splits images with the same size during the whole training process, which means it is difficult to exploit multi-granularity regions. In this work, we apply a jigsaw puzzle generator to restrict the granularity of learned regions at each training step.

\subsection{Progressive Training}
Progressive training methodology was originally proposed for generative adversarial networks~\cite{karras2017progressive}, where it started with low-resolution images, and then progressively increased the resolution by adding layers to the networks. Instead of learning the information from all the scales, this strategy allows the network to discover large-scale structure of the image distribution and then shift attention to increasingly ﬁner scale details.  Recently, progressive training strategy has been widely utilized for generation tasks \cite{karras2019style,shaham2019singan,wang2018fully,ahn2018image}, since it can simplify the information propagation within the network by intermediate supervision. 

For FGVC, the fusion of multi-granularity information is critical to the model performance. In this work, we adopt the idea of progressive training to design a single network that can learn these information with a series of training stages. The input images are firstly split into small patches to train a low-level layers of model. Then the number of patches are progressively increased and the corresponding layers high-level lays have been added and trained, correspondingly. Most of the existing work with progressive training are focusing on the task of sample generation. To the best of our knowledge, it has not been attempted earlier for the task of FGVC.

\section{Approach}

\begin{figure}[!t]
\centering
\includegraphics[width=10cm]{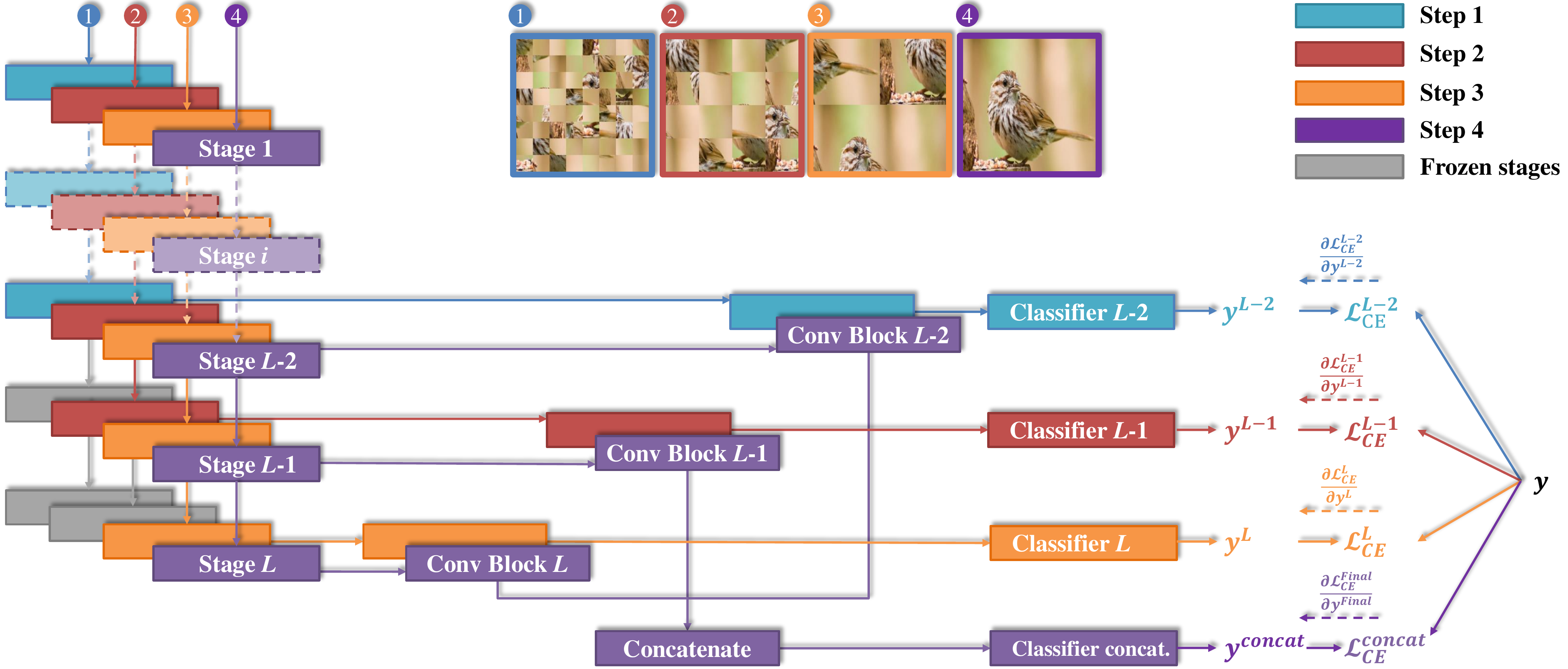}
\caption{
The training procedure of the progressive training which consists of $S+1$ steps at each iteration (Here $S=3$ for explanation). The $Conv$ $Block$ represents the combination of two convolution layers with and max pooling layer, and $Classifier$ represent two fully connected layers with a softmax layer at the end. At each iteration, the training data are augmented by the jigsaw generator and sequentially input into the network by $S+1$ steps. In our training process, the hyper-parameter $n$ is $2^{L-l+1}$ for the $l^{th}$ stage. At each step, the output from the corresponding classifier will be used for loss computation and parameter updating.
}
\label{fig:network_train}
\end{figure}

In this section, we present our proposed Progressive Multi-Granularity (PMG) training framework. As shown in Figure \ref{fig:network_train}, to address the large intra-class variations, we encourage the model to learn stable fine-grained information in the shallower layers and gradually shift attention to the learning of abstract information of large granularity level in the deeper layers as training progresses.

\subsection{Network Architecture}

Our network design is generic and could be implemented on the top of any state-of-the-art backbone feature extractor, like Resnet \cite{he2016deep}. Let us ${F}$ be our backbone feature extractor, which has $L$ stages. The output feature-map from any intermediate stages is represented as ${F}^{l} \in  \mathbb{R}^{H_{l} \times W_{l} \times C_{l}}$, where $H_{l}$, $W_{l}$, $C_{l}$ are the height, width and number of channels of the feature map at $l$-th stage, and $l=\{1,2, ...,~L\}$. Here, our objective is to impose classification loss on the feature-map extracted at different intermediate stages. Hence, in addition to ${F}$, we introduce convolution block ${H}_{conv}^{l}$ that takes $l$-th intermediate stage output ${F}^{l}$ as input and reduces it to a vector representation ${V}^{l} = {H}_{conv}^{l}({F}^{l})$. Thereafter, a classification module ${H}_{class}^{l}$ consisting of two fully-connected stage with Batchnorm \cite{ioffe2015batch} and Elu\cite{clevert2015fast} non-linearity, corresponding to $l$-th stage, predicts the probability distribution over the classes as $y^{l} = {H}_{class}^{l}({V}^{l})$. Here, we consider last $S$ stages: $l = L, L-1, \dotsc, L-S+1$. Finally, we concatenate the output from last three stages as 
\begin{equation}
{V}^{concat} = \text{concat}[{V}^{L-S+1}, \dotsc, {V}^{L-1}, {V}^{L}]  
\end{equation}
 This is followed by an additional classification module  $y^{concat} = {H}_{class}^{concat}({V}^{concat})$




\subsection{Progressive Training}

We adopt progressive training where we train the low stage first and then progressively add new stages for training. Since the receptive field and representation ability of low stage is limited, the network will be forced to first exploit discriminative information from local details ({\em i.e.} object textures). Compared to training the whole network directly, this increment nature allows the model to locate discriminative information from local details to global structures when the features are gradually sent into higher stages, instead of learning all the granularities simultaneously.

For the training of the outputs from each stages and the output from the concatenated features, we adopt cross entropy (CE) $\mathscr{L}_{CE}$ between ground truth label $y$ and prediction probability distribution for loss computation as

\begin{equation}
    \mathscr{L}_{CE}(y^l, y) = - \sum_{i=1}^{m} y^l_i \times log(y^l_i). \label{eq:no0}
\end{equation}

and

\begin{equation}
    \mathscr{L}_{CE}(y^{concat}, y) = - \sum_{i=1}^{m} y^{concat}_i \times log(y^{concat}_i). \label{eq:no1}
\end{equation}

At each iteration, a batch of data $d$ will be used for $S+1$ steps, and we only train one stage's output at each step in series. It needs to be clear that all parameters are used in the current prediction will be optimized, even they may have been updated in the previous steps, and this can help each stage in the model work together.

\subsection{Jigsaw Puzzle Generator}

Jigsaw Puzzle solving \cite{wei2019iterative} has been found to be suitable for self-supervised task in representation learning. On the contrary, we borrow the notion of Jigsaw Puzzle to generate input images for different steps of progressive training. The objective is to devise
different granularity regions and force the model to learn information specific to the corresponding granularity level at each training step. Given an input image $d\in R^{3\times W\times H}$, we equally split it into $n \times n$ patches which have $3\times\frac{W}{n}\times\frac{H}{n}$ dimensions. $W$ and $H$ should be integral multiples of $n$, respectively. Then, the patches are shuffled randomly and merged together into a new image $P(d, n)$. Here, the granularities of patches are controlled by the hyper-parameter $n$.

Regarding the choice of hyper-parameter $n$ for each stage, two conditions needs to be satisfied: (i) the size of the patches should be smaller than the receptive field of the corresponding stage, otherwise, the performance of the jigsaw puzzle generator will be reduced; (ii) the patch size should increase proportionately with the increase of the receptive fields of the stages. Usually, the receptive field of each stage is approximately double than that of the last stage. Hence, we set $n$ as $2^{L-l+1}$ for the $l^{th}$ stage's output.

During training, a batch $d$ of training data will first be augmented to several jigsaw puzzle generator-processed batches, obtaining $P(d, n)$. All the jigsaw puzzle generator-processed batches share the same label $y$. Then, for the $l^{th}$ stage's output ${y}^{l}$, we input the batch $P(d, n),n=2^{L-l+1}$, and optimize all the parameters used in this propagation. Figure \ref{fig:network_train} illustrates the training procedure step by step.

It should be clarified that the jigsaw puzzle generator cannot always guarantee the completeness of all the parts which are smaller than the size of the patch. Although there could exist some parts which are smaller than the patch size, those still have chances of getting split. However, it should not be a bad news for model training, since we adopt random cropping which is a standard data augmentation strategy before the jigsaw puzzle generator and leads to the result that patches are different compared with those of previous iterations. Small discriminative parts, which are split at this iteration due to the jigsaw puzzle generator, will not be always split in other iterations. Hence, it brings an additional advantage of forcing our model to find more discriminative parts at the specific granularity level.

\subsection{Inference}

At the inference step, we merely input the original images into the trained model and the jigsaw puzzle generator is unnecessary. If we only use ${y}^{concat}$ for prediction, the FC layers for the other three stages can be removed which leads to less computational budget. In this case, the final result $C_1$ can be expressed as
\begin{equation}
    C_1 = argmax({y}^{concat}). \label{eq:no4}
\end{equation}

However, the prediction from a single stage based on information of a specific granularity is unique and complementary, which leads to a better performance when we combine all outputs together with equal weights. The multi-output combined prediction $C_2$ which can be written as

\begin{equation}
    C_2 = argmax(\sum_{l=L-S+1}^L {y}^{l} + {y}^{concat}).\label{eq:no3}
\end{equation}

Hence, both the prediction of ${y}^{concat}$ and multi-output combined prediction can be obtained in our model. In addition, although all predictions are complementary for final result, ${y}^{concat}$ is enough for those objects whose shapes are relatively smooth, for example, cars. More details of experiments could be found in Section \ref{sec:experiments}.

\section{Experiment Results and Discussion} \label{sec:experiments}

In this section, we evaluate the performance of the proposed method on three ﬁne-grained image classiﬁcation datasets: Caltech UCSD-Birds (CUB) \cite{wah2011caltech}, Stanford Cars (CAR) \cite{krause20133d}, and FGVC-Aircraft (AIR) \cite{maji2013fine}. Firstly, the implementation details are introduced in Section \ref{ssec:implementation}. Subsequently, the classiﬁcation accuracy comparisons with other state-of-the-art methods are then provided in Section \ref{ssec:sota}. In order to illustrate the advantages of different components and design choices in our method, a comprehensive ablation study and a visualization are provided in Section \ref{ssec:ablation} and \ref{ssec:visualization}.

\subsection{Implementation Details}\label{ssec:implementation}
 We perform all experiments using PyTorch \cite{paszke2017automatic} with version higher than 1.3 over a cluster of GTX 2080 GPUs. The proposed method is evaluated on the widely used backbone networks: VGG16 \cite{simonyan2014very} and ResNet50 \cite{he2016deep}, which means the total number of stages $L=5$. For the best performance, we set $S=3$, $\alpha=1$, and $\beta=2$. The category labels of the images are the only annotations used for training. The input images are resized to a ﬁxed size of $550 \times 550$ and randomly cropped into $448 \times 448$, and random horizontal ﬂip is applied for data augmentation when we train the model. During testing, The input images are resized to a ﬁxed size of $550 \times 550$ and cropped from center into $448 \times 448$. All the above settings are standard in the literatures. 
 
 We use stochastic gradient descent (SGD) optimizer and batch normalization as the regularizer. Meanwhile, the learning rates of the convolution layers and the FC layers, respectively, which are newly added by us are initialized as 0.002 and reduced by following the cosine annealing schedule \cite{loshchilov2016sgdr} during training. The learning rates of the pre-trained convolution layers are maintained as 1/10 of those of the newly added layers. For all the aforementioned models, we train them for up to 300 epochs with batch size as 16 and used a weight decay as 0.0005 and a momentum as 0.9.

\subsection{Comparisons with State-of-the-Art Methods}\label{ssec:sota}
The comparisons of our method with other state-of-the-art methods on CUB-200-2011, Stanford Cars, and FGVC-Aircraft are presented in Table~\ref{table:comparison}. Both the accuracy of ${y}^{concat}$ and the combined accuracy of all four outputs are listed.

\setlength{\tabcolsep}{4pt}
\begin{table}[!t]
\begin{center}
\caption{comparison results with state-of-the-art methods.}
\label{table:comparison}
\begin{tabular}{lcccc}
\hline\noalign{\smallskip}
Method & Base Model & CUB (\%) & CAR (\%) & AIR (\%)\\
\noalign{\smallskip}
\hline
\noalign{\smallskip}
FT VGG (CVPR18) \cite{wang2018learning} & VGG16 & 77.8 & 84.9 & 84.8\\
FT ResNet (CVPR18) \cite{wang2018learning} & ResNet50 & 84.1 & 91.7 & 88.5\\
B-CNN (ICCV15) \cite{lin2015bilinear} & VGG16 & 84.1 & 91.3 & 84.1\\
KP (CVPR17) \cite{cui2017kernel} & VGG16 & 86.2 & 92.4 & 86.9\\
RA-CNN (ICCV17) \cite{fu2017look} & VGG19 & 85.3 & 92.5 & -\\
MA-CNN (ICCV17) \cite{zheng2017learning} & VGG19 & 86.5 & 92.8 & 89.9\\
PC (ECCV18) \cite{dubey2018pairwise} & DenseNet161 & 86.9 & 92.9 & 89.2\\
DFL (CVPR18) \cite{wang2018learning} & ResNet50 & 87.4 & 93.1 & 91.7\\
NTS-Net (ECCV18) \cite{yang2018learning} & ResNet50 & 87.5 & 93.9 & 91.4\\
MC-Loss (TIP20) \cite{chang2019the} & ResNet50 & 87.3 & 93.7 & 92.6\\
DCL (CVPR19) \cite{chen2019destruction} & ResNet50 & 87.8 & 94.5 & \underline{93.0}\\
MGE-CNN (ICCV19) \cite{zhang2019learning} & ResNet50 & 88.5 & 93.9 & -\\
S3N (ICCV19) \cite{ding2019selective} & ResNet50 & 88.5 & 94.7 & 92.8\\
Stacked LSTM (CVPR19) \cite{ge2019weakly} & ResNet50 & {\bf 90.4} & - & -\\
\hline
PMG & VGG16 & 88.2 & 94.2 & 92.4\\
PMG (Combined Accuracy) & VGG16 & 88.8 & 94.3 & 92.7\\
PMG & ResNet50 & 88.9 & \underline{95.0} & 92.8\\
PMG (Combined Accuracy) & ResNet50 & \underline{89.6} & {\bf 95.1} & {\bf 93.4}\\
\hline
\end{tabular}
\end{center}
\end{table}
\setlength{\tabcolsep}{1.4pt}

\subsubsection{CUB-200-2011}

We achieve competitive result on this dataset in a much easier experimental procedure, since only one network and one propagation are needed during testing. Our method outperform RA-CNN\cite{fu2017look} and MGE-CNN \cite{zhang2019learning} by 4.3\% and 1.1\%, even though they build several different networks to learn information of various granularities. They train the classification of each network separately and then combine their information for testing, which proofs our advantage of exploiting multi-granularity information gradually in one network. Besides, even Stacked LSTM \cite{ge2019weakly} better performance than our method, it is a two phase algorithm that requires Mask-RCNN \cite{he2017mask} and CPF to offer complementary object parts and then use bi-directional LSTM \cite{hochreiter1997long} for classification, which leads to more inference time and computation budget.

\subsubsection{Stanford Cars}

Our method achieves state-of-the-art performance with Resnet50 as the base model. Since the cars at Stanford Cars dataset are much more rigid and performance of ${y}^{concat}$ is good enough, the improvement of combining multi-stage outputs is not obvious. The result of our method surpasses PC \cite{dubey2018pairwise} even it improves its performance by adopting more advanced backbone network {\em i.e.} DenseNet161. For MA-CNN \cite{zheng2017learning} and NTS-Net \cite{yang2018learning} which first locate several different discriminative parts and then combine features extracted from them for final classification, we outperform them by large margins of 2.3\% and 1.2\%.

\subsubsection{FGVC-Aircraft}

On this task, the multi-stage outputs combined result of our method also achieves State-of-the-Art performance. Although S3N \cite{ding2019selective} find both discriminative part and complementary part for feature extraction and apply additional inhomogeneous transform to highlight these parts, we still outperform it by 0.6\% with the same backbone network ResNet50, and show competitive result even when we adopt VGG16 as the base network.

\subsection{Ablation Study}\label{ssec:ablation}

We conduct ablation studies to understand the effectiveness of the progressive training strategy and the jigsaw puzzle generator. We choose CUB-200-2011 dataset for experiments and ResNet50 as the backbone network, which means the total number of stages $L$ is $5$. We first design different runs with the number of stages used for output $S$ increasing from $1$ to $5$ and no jigsaw puzzle generator, as shown in Table \ref{table:5stage_woJPG}. The ${y}^{concat}$ is kept for all runs and number of steps is $S+1$. It is clear that the increasing of $S$ boosts the model performance significantly when $S<4$. However, we also notice the accuracy starts to decrease when $S$ become $4$. The possible reason is that low stage layers are mainly focus on class-irrelevant features, but the additional supervision will force it to find class-relevant information and then affect the overall performance.

In Table \ref{table:5stage_wJPG}, we report the results of our method with assistance of the jigsaw puzzle generator. The hyper-parameter $n$ of the jigsaw puzzle generator for $l^{th}$ stage follows the pattern that $n=2^{L-l+1}$. Compared with results in Table \ref{table:5stage_wJPG}, the jigsaw puzzle generator improves the model performance on the basis of progressive training when $S<4$. When $S=4$, the model with the jigsaw puzzle generator does not show any advantages, and when $S=5$ the jigsaw puzzle generator lowers the model performance. This is because when $n>8$ the split patches are too small to keep meaningful information, which confuses the model training.

According to the above analysis, progressive training are beneficial for fine-grained classification task when we choose appropriate $S$. In such a case, the jigsaw puzzle generator can further improve the performance.

\setlength{\tabcolsep}{4pt}
\begin{table}[!t]
\begin{center}
\caption{The accuracy and combined accuracy of proposed method by using different hyper-parameters $s$ without the assistance of jigsaw puzzle generator}
\label{table:5stage_woJPG}
\begin{tabular}{lcc}
\hline\noalign{\smallskip}
S,n & Accuracy (\%) & Combined Accuracy (\%)\\
\noalign{\smallskip}
\hline
\noalign{\smallskip}
1,\{1,1\} & 86.3 & 86.5\\
2,\{1,1,1\} & 87.6 & 88.0\\
3,\{1,1,1,1\} & {\bf 88.3} & {\bf 88.7}\\
4,\{1,1,1,1,1\} & 87.8 & 88.5\\
5,\{1,1,1,1,1,1\} & 87.7 & 88.3\\
\hline
\end{tabular}
\end{center}
\end{table}
\setlength{\tabcolsep}{1.4pt}

\setlength{\tabcolsep}{4pt}
\begin{table}[!t]
\begin{center}
\caption{The accuracy and combined accuracy of proposed method by using different hyper-parameters $s$ and the corresponding set of $n$}
\label{table:5stage_wJPG}
\begin{tabular}{lcc}
\hline\noalign{\smallskip}
S,n & Accuracy (\%) & Combined Accuracy (\%)\\
\noalign{\smallskip}
\hline
\noalign{\smallskip}
1,\{2,1\} & 86.9 & 86.9\\
2,\{4,2,1\} & 88.5 & 88.7\\
3,\{8,4,2,1\} & {\bf 88.9} & {\bf 89.6}\\
4,\{16,8,4,2,1\} & 88.0 & 88.5\\
5,\{32,16,8,4,2,1\} & 87.2 & 87.7\\
\hline
\end{tabular}
\end{center}
\end{table}
\setlength{\tabcolsep}{1.4pt}

\subsection{Visualization}\label{ssec:visualization}

\begin{figure}[!t]
\centering
  \includegraphics[width=10cm]{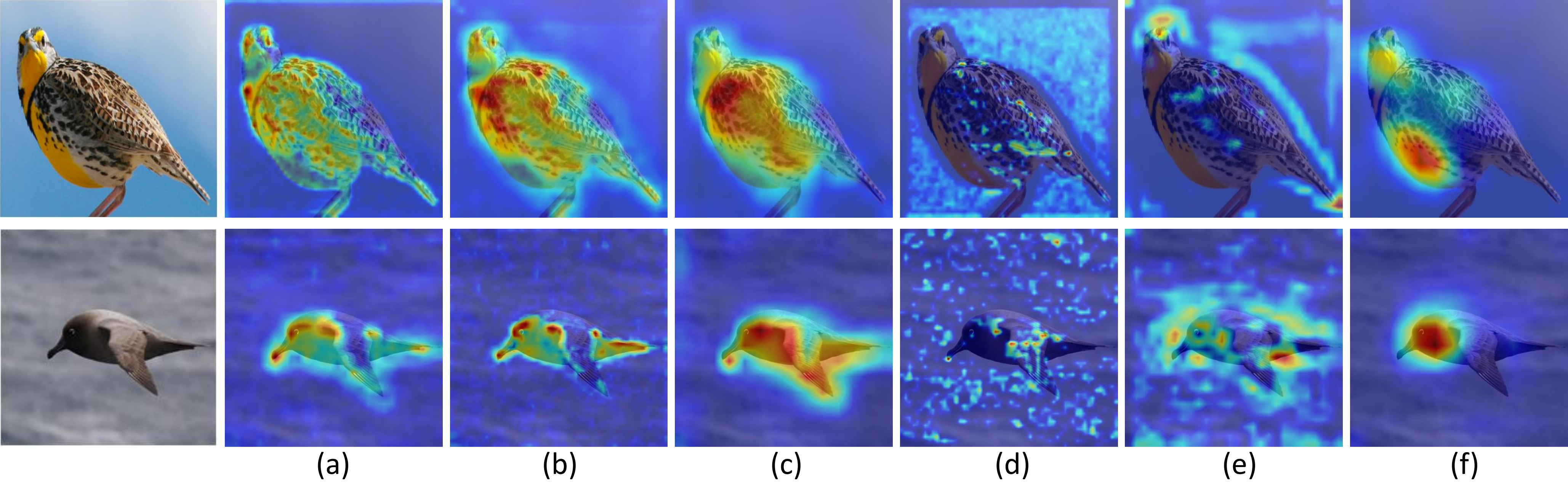}
  \caption{Activation map of selected results on the CUB dataset with the Resnet50 as the base model. Columns (a)-(c) are visualization of the convolution layer from the third to the fifth stage of our model. Columns (d)-(e) are visualization of the convolution layer from the third to the fifth stage of the baseline model.}
  \label{fig:visualization}
\end{figure}

In order to illustrate the advantages of the proposed method, we apply the Grad-CAM to implement the visualization for last three stages' convolution layer of both our method and baseline model. Columns (a)-(c) in Figure \ref{fig:visualization} are visualization of the convolution layers from the third to the fifth stage of our model's backbone, which are supervised by images generated by jigsaw puzzle generator with $n=\{8,4,2\}$ sequentially. It is clear in column (a) that the model concentrates on discriminative parts of small granularity at the third stage like bird eyes and small pattern or texture of birds' feather. And when it comes to column (c), the fifth stage of the model pays attention to parts of larger granularity. The visualization result demonstrates that our model truly gives predictions based on discriminative parts of small granularity to large granularity gradually.

When compared with the activation map of the baseline model, our model shows more meaningful concentration on the target object, while the baseline model only shows the correct attention at the last stage. This difference indicates that the intermediate supervision of progressive training can help the model locate useful information at earlier stages. Besides, we find the baseline model usually only concentrates on one or two parts of the object at the last stage where it makes prediction. However, the attention regions of our method nearly cover the whole object at each stage, which indicates that images generated by the jigsaw puzzle generator can forcing the model to learn more discriminative parts at each granularity level.

\section{Conclusions}

In this paper apply progressive training strategy into fine-grained classification tasks and propose a novel framework named Progressive
Multi-Granularity (PMG) Training with two main components: (i) a novel training strategy that fuses multi-granularity features in a progressive manner, and (ii) a simple jigsaw puzzle generator to form images contain information of different granularity levels. Our method can be trained end-to-end without other manual annotations except category labels, and only needs one network with one propagation during testing. We conduct experiments on three widely used fine-grained datasets and obtain state-of-the-art performance on two of them and a competitive result on the other one, which demonstrate the effectiveness of our method.

\clearpage
\bibliographystyle{splncs04}
\bibliography{egbib}
\end{document}